\documentclass[sigconf]{acmart}

\newcommand{\eg}{\emph{e.g.,}}
\newcommand{\ie}{\emph{i.e.,}}
\newcommand{\etal}{\emph{et~al.}}
\usepackage{fancyhdr}
\usepackage[normalem]{ulem}
\usepackage{adjustbox}
\usepackage[T1]{fontenc}
\usepackage{subcaption}

\usepackage{amsmath}
\usepackage{amssymb}
\usepackage{amsthm}
\usepackage[para,online,flushleft]{threeparttable}
\usepackage{tikz}

\renewcommand\footnotetextcopyrightpermission[1]{} 
\begin{document}
\title{Full-stack Optimization for Accelerating CNNs\\with FPGA Validation}

\author{Bradley McDanel$^*$}\thanks{*Equal Contribution.}
\affiliation{%
  \institution{Harvard University}
}
\email{mcdanel@fas.harvard.edu}

\author{Sai Qian Zhang$^*$}
\affiliation{%
  \institution{Harvard University}
}
\email{zhangs@g.harvard.edu}

\author{H.T. Kung}
\affiliation{%
  \institution{Harvard University}
}
\email{kung@harvard.edu}

\author{Xin Dong}
\affiliation{%
  \institution{Harvard University}
}
\email{xindong@g.harvard.edu}

\begin{abstract}
We present a full-stack optimization framework for accelerating inference of CNNs (Convolutional Neural Networks) and validate the approach with field-programmable gate arrays (FPGA) implementations. By jointly optimizing CNN models, computing architectures, and hardware implementations, our full-stack approach achieves unprecedented performance in the trade-off space characterized by inference latency, energy efficiency, hardware utilization and inference accuracy. As a validation vehicle, we have implemented a 170MHz FPGA inference chip achieving 2.28ms latency for the ImageNet benchmark. The achieved latency is among the lowest reported in the literature while achieving comparable accuracy. However, our chip shines in that it has 9x higher energy efficiency compared to other implementations achieving comparable latency. A highlight of our full-stack approach which attributes to the achieved high energy efficiency is an efficient Selector-Accumulator (SAC) architecture for implementing the multiplier-accumulator (MAC) operation present in any digital CNN hardware. For instance, compared to a FPGA implementation for a traditional 8-bit MAC, SAC substantially reduces required hardware resources (4.85x fewer Look-up Tables) and power consumption (2.48x).
\end{abstract}

\maketitle

\begin{tikzpicture}[overlay, remember picture]
\path (current page.north east) ++(-1.22,-1) node[below left] {Final version will appear in International};
\path (current page.north east) ++(-1,-1.4) node[below left] {Conference on Supercomputing (ICS) 2019.};
\end{tikzpicture}

\section{Introduction}
Due to the widespread success of Convolutional Neural Networks (CNNs) across a variety of domains, there have been extraordinary research and development efforts placed on improving inference latency, energy efficiency, and accuracy of these networks. Generally, these research efforts can be viewed from two distinct perspectives: (1) machine learning practitioners who focus on reducing the complexity of CNNs through more efficient convolution operations~\cite{wu2017shift}, weight and activation quantization~\cite{jacob2017quantization}, and weight pruning~\cite{han2015deep} and (2) hardware architecture experts who design and build CNN accelerators with minimal power consumption and I/O cost~\cite{du2015shidiannao,jouppi2017datacenter,zhang2016cambricon,wang2017chain}. 

\begin{figure}
    \centering
    \includegraphics[width=\columnwidth]{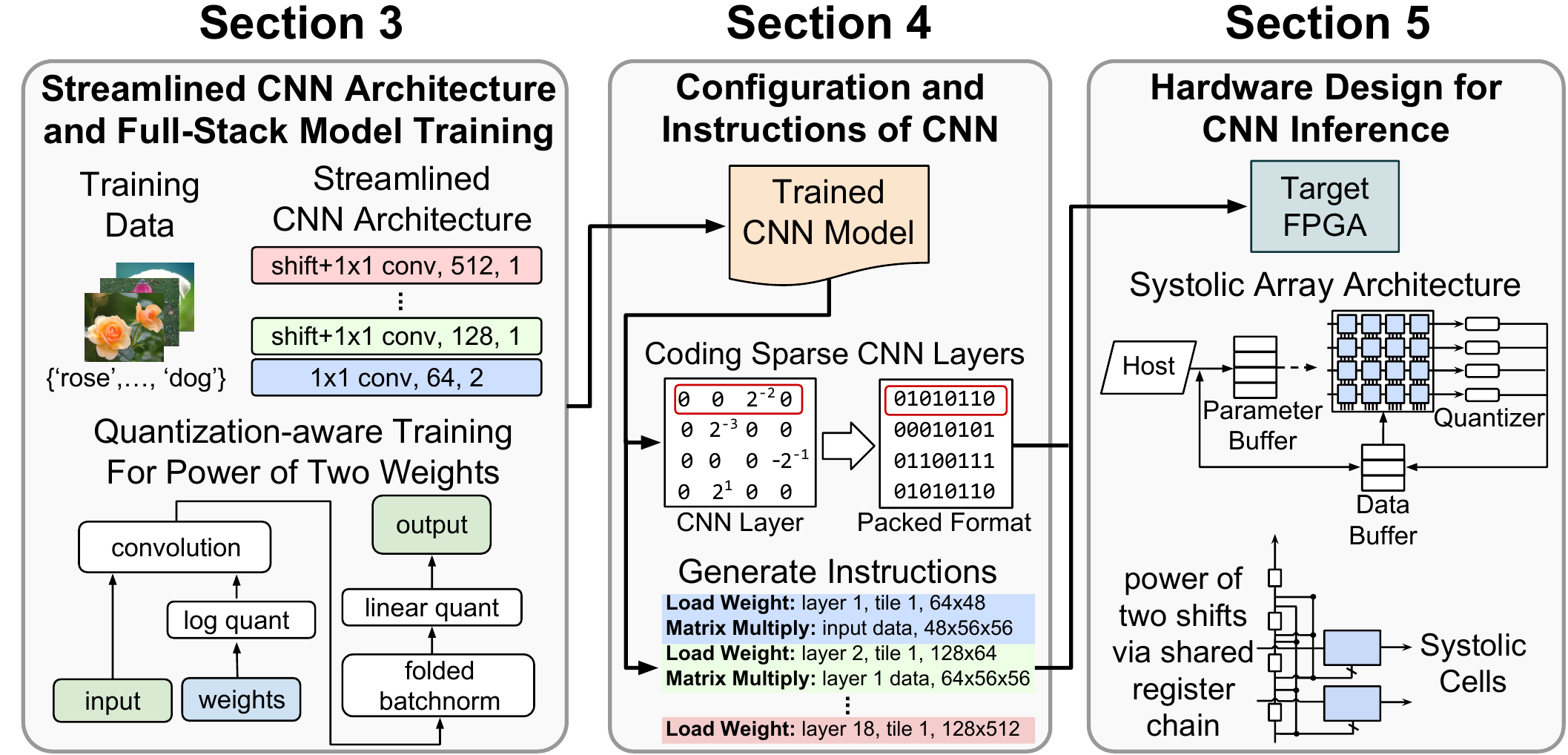}
    \caption{An overview of the proposed full-stack optimization framework for accelerating inference of sparse CNNs. Section 3 details CNN training which includes constraints to match the proposed computing architecture. Section 4 described the process of converting a trained sparse CNN into a packed representation and how FPGA instructions are generated for each layer in the CNN. Section 5 outlines the proposed architecture including the use of multiplication-free selector-accumulator (SAC) based systolic cells which are used to perform inference for all CNN layers on the FPGA.}
    \label{fig:overview}
\end{figure}

However, approaching the problem from only one of these two viewpoints can lead to suboptimal solutions. For instance, as discussed in Section~\ref{sec:quantization}, many low-precision weight quantization methods omit certain significant cost factors in an end-to-end implementation such as using full-precision weights and data for the first and last layers~\cite{zhou2017incremental,cai2017deep} or full-precision batch normalization~\cite{ioffe2015batch} as in ~\cite{rastegari2016xnor}. On the other side, most CNN accelerators are designed to support some target CNNs (\eg~AlexNet~\cite{krizhevsky2012imagenet} and VGG-16~\cite{simonyan2014very}) at 8-bit or 16-bit precision for weights or data~\cite{gupta2015deep,dettmers20158}. Therefore, these accelerators generally are not directly applicable to many of the recent advances in CNN design including efficient CNN structures (using,~\eg~separable filters~\cite{howard2017mobilenets}), weight pruning (using,~\eg~Lasso~\cite{tibshirani1996regression}), and low-precision quantization. 

To address this disparity, in this paper we propose a \textit{full-stack optimization} framework, where the design of the CNN model is \textit{jointly} optimized with the computing architectures and circuit implementations on which it will run. Figure~\ref{fig:overview} provides an overview of the proposed method in three stages, covered in three sections. Section~\ref{sec:training} describes the training stage, which uses a hardware-aware quantization graph to facilitate training a CNN which can be directly implemented on an FPGA without any additional overhead. 

\begin{figure}
    \centering
    \includegraphics[width=0.85\columnwidth]{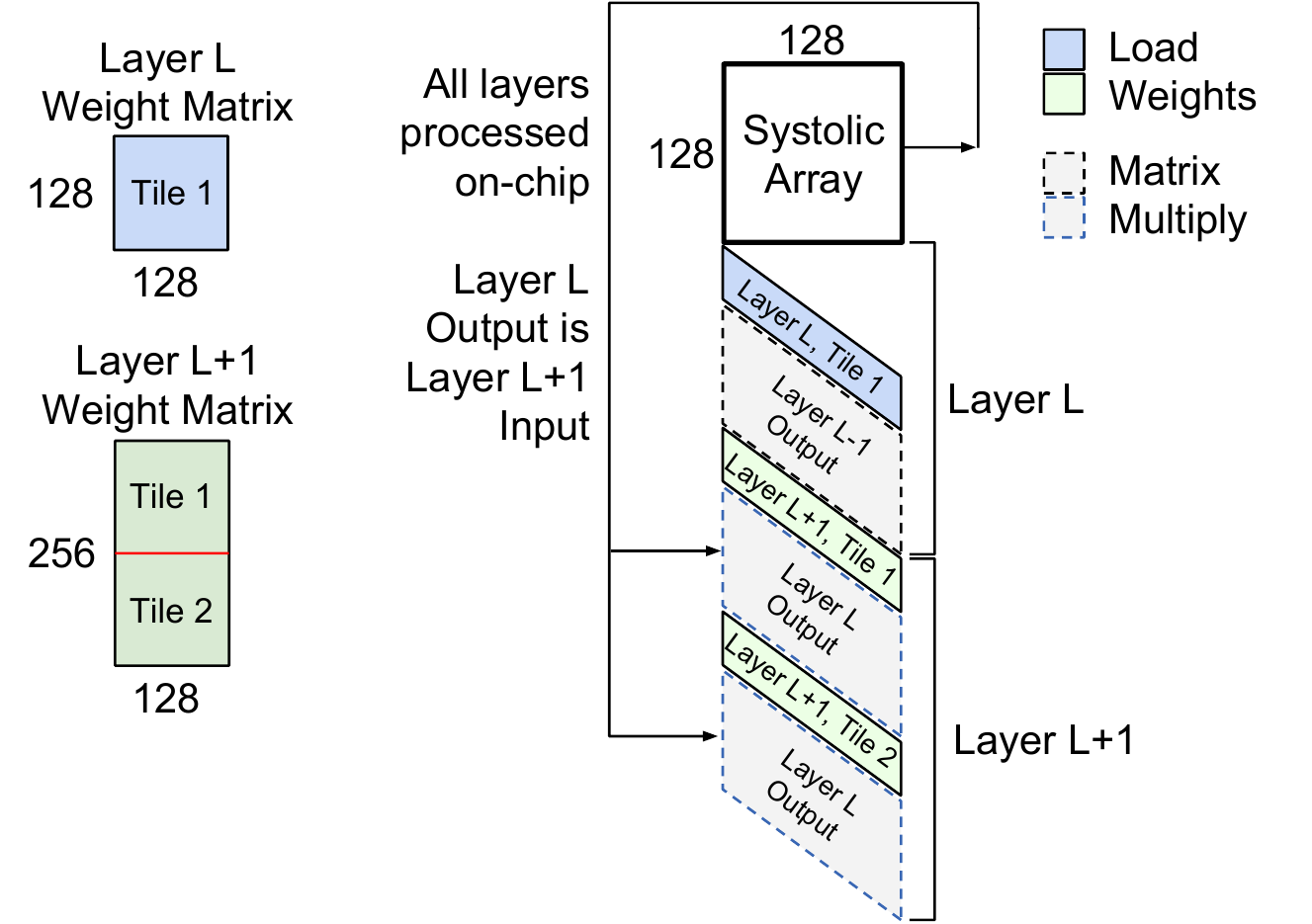}
    \caption{CNN inference for two consecutive layers (layer L and L + 1) using a single 128$\times$128 systolic array similar to the systolic array implemented on the FPGA reported in this paper. The systolic array alternatively executes load weight and matrix multiply instructions for all tiles in a layer (six instructions in total for this example; see Section~\ref{sec:inst-generation}).}
    \label{fig:tiling}
\end{figure}

Section~\ref{sec:cnn-conversion} describes the process of  generating instructions to perform inference given both the trained CNN and also a systolic array of a fixed size implemented on the target FPGA. It also covers how the trained sparse and quantized CNN is coded for efficient use of FPGA memory. Section~\ref{sec:fpga-inference} depicts the bit-serial systolic array architecture which includes the use of multiplication-free sparse systolic cells, based on the Selector-Accumulator (SAC) architecture for the multiplier-accumulation (MAC) operation, for efficient inference and can leverage irregular sparsity in a CNN model. Figure~\ref{fig:tiling} depicts how a given systolic array implemented on a FPGA carries out a CNN inference by reusing the array across the CNN layers. Note that for systolic array synchronization, items input to and output from the array are properly skewed, as shown in the figure. When a layer has more filters than the systolic array can handle, we partition the layer into \textit{vertical tiles} across filters, as shown on the left of the figure, and reuse the systolic array across these tiles. When a layer has more channels than the systolic array can handle, we partition the layer into \textit{horizontal tiles} across channels (these horizontal tiles are not shown in the figure). In this paper, with column combining~\cite{kungpacking18}, the 128x64 systolic array implemented on our FPGA is large enough to handle all channels in each layer of evaluation CNN models. Thus we do not use horizontal tiles.

The novel techniques of the paper are as follows:
\begin{itemize}
    \item A \textbf{full-stack optimization framework} where the hardware structure is used to inform the CNN structure via training, and vice versa.
    \item \textbf{Selector-accumulator} (SAC) which provide efficient\\multiplication-free inference. We replace traditional \\multiplier-accumulator (MAC) hardware with inexpensive SAC circuits facilitated by simple shared register chains (Section~\ref{sec:mx-design}).
    \item A \textbf{systolic array building block} for low-precision CNNs (Section~\ref{sec:systolic-array-system}) which uses shared register chains for two purposes: propagating data into adjacent cells and performing multiplication with power of two weights. Systolic arrays are used as example of processor arrays (our register chain design may extend to other array architectures).
    \item A \textbf{streamlined CNN structure} which achieves competitive performance on ImageNet in the mobile setting using only 1$\times$1 convolution without residual connections (Section~\ref{sec:cnn-structure}).
    \item \textbf{Input reshaping} which allows parallel input of input image into the systolic array (Section~\ref{sec:reshape-input}).
\end{itemize}

Leveraging all these advances into a single system is challenging and one of the main accomplishments of this work. We have built an efficient CNN inference engine on a FPGA (Xilinx VC707 evaluation board) and have validated its correctness by checking the output against our Python simulator's output. All the timing and power consumption results reported in this paper are based on the actual measurements obtained from this FPGA implementation. Our FPGA design is composed almost entirely of Look-up Tables (LUTs). We use DSPs only to implement the final fully connected layer. We believe our design provides a useful base for future ASIC implementation. Additionally, the concepts and architecture presented in this paper could scale across multiple components in a distributed setting to form larger systems for deep learning.

Links to our CNN training code (using PyTorch~\cite{paszke2017automatic}), python code which converts a trained CNN into a packed representation for the FPGA, and Verilog code for FPGA implementation is available at \url{https://goo.gl/8i9aJp}.

\section{Background and Related Work}
\label{sec:background}
In this section, we first summarize recent FPGA-based CNN accelerators which we compare against in Section~\ref{sec:eval}. Then, we review advances in efficient CNNs we use as a starting point for our approach.

\subsection{FPGA Accelerators for CNNs}
In recent years, numerous FPGA designs for CNN inference have been proposed (generally targeting prominent networks such as LeNet-5~\cite{lecun1998gradient}, AlexNet~\cite{krizhevsky2012imagenet}, and VGG-16~\cite{simonyan2014very}) with the key objective of designing a system with low latency and high energy efficiency. A common strategy deployed by these designs is to minimize the degree of weight and data movement, especially from off-chip memory, as they add significant overhead in terms of both latency and power consumption. 

One approach to minimizing data movement is layer fusion, where multiple CNN layers are processed at the same time in a pipelined manner to allow for instant use of intermediate data without external memory accesses~\cite{alwani2016fused,li2016high,xiao2017exploring}. Another approach, used for 3$\times$3 or larger convolutional filters, is to determine the ordering of inference computation which minimizes the number of partial sums that must be stored~\cite{ma2017optimizing,zhang2018dnnbuilder}. Since our streamlined CNN architecture (Section~\ref{sec:training}) uses only 1$\times$1 filters, convolution is reduced to matrix multiplication, which can be efficiently implemented using systolic arrays. Additionally, different computation strategies are often taken for the first layer~\cite{xilinxtech18}, as it has only three input channels in the case of RGB images and final fully connected layer~\cite{qiu2016going}, where there are significantly more weights than data. In this work, we propose to use the same systolic array building block for efficient implementations of all layers in a CNN by using various full-stack optimization techniques such as input reshaping discussed in Section~\ref{sec:reshape-input}.

\subsection{Efficient CNN Structures}
\label{sec:cnn-background}

Since VGG-16~\cite{simonyan2014very} was introduced in 2014, there has been a general trend towards designing deeper CNNs through the use of residual connections (ResNets\cite{he2016deep}) and concatenative connections (DenseNet~\cite{huang2017densely}) as deeper networks tend to achieve higher classification accuracy for benchmark datasets such as ImageNet~\cite{deng2009imagenet}. However, as pointed out in Table 2 of the original ResNet paper~\cite{he2016deep}, residual connections appear to add little improvement in classification accuracy to a shallower (18 layer) CNN. Based on these observations, we have chosen to use a shallower CNN (19 layers) without any residual or concatenative connections, which we outline in Section~\ref{sec:cnn-structure}. In our evaluation (Section~\ref{sec:residual-comp}) we show that for this shallower CNN, the exclusion of additional connections has minimal impact on classification accuracy while significantly simplifying our hardware implementation and improving its efficiency.

Additionally, several alternatives to standard convolution have recently been proposed to reduce the computation cost. Depthwise separable convolution~\cite{chollet2016xception} dramatically reduces the number weights and operations by separating a standard convolution layer into two smaller layers: a depthwise layer that only utilize neighboring pixels within each input channel and a pointwise layer which operates across all channels but does not use neighboring pixels within a channel (\ie~it only uses 1$\times$1 filters). Wu~\etal~showed that a channel shift operation can be used to replace the depthwise layer without significant impact on classification accuracy~\cite{wu2017shift}. As described in Section~\ref{sec:cnn-structure}, our proposed CNN use this channel shift operation immediately preceding a 1$\times$1 convolution layer.

\subsection{Weight and Data Quantization}
\label{sec:quantization}

Several methods have been proposed to quantize the CNN weights after training, using 16-bit~\cite{gupta2015deep} and 8-bit~\cite{dettmers20158} fixed-point representations, without dramatically impacting classification accuracy. More recently, low-precision quantization methods (\ie~1-4 bits) such as binary~\cite{courbariaux2015binaryconnect,hu2018hashing} and ternary quantization~\cite{zhu2016trained,zhou2017incremental,wang2017fixed} methods have also been studied, which to smaller models may incur some cost to classification accuracy. Generally, for these low-precision approaches, training is still performed using full-precision weights, but the training graph is modified to include quantization operations in order to match the fixed-point arithmetic used at inference. In this paper, log quantization~\cite{zhou2017incremental} is adopted for weights, with each quantization point being a power of two. This allows for significantly more efficient inference, as fixed-point multiplication is replaced with bit shift operations corresponding the power of two weight, as discussed in Section~\ref{sec:fpga-inference}.

In addition to weight quantization, there are many quantization methods for activation data output from each CNN layer~\cite{rastegari2016xnor,zhou2016dorefa,cai2017deep,zhou2017incremental,choi2018pact}. Data quantization reduces the cost of memory access for these intermediate output between layers in a CNN and also the computation cost of inference. However, it has been shown that low-precision quantization of activation (\ie~1-4 bits) leads to a significantly larger degradation in classification accuracy compared to weight quantization~\cite{courbariaux2016binarized,liu2018bi}. Due to these considerations, we use 8-bit linear quantization for data in this paper and focus on an efficient implementation of multiplication-free computations with 8-bit data.

Additionally, we note that the majority of proposed methods for low precision weights and data omit two details which are critical for efficient end-to-end system performance. First, works in this area often treat the first and last layers in a special manner by keeping the weights and data full-precision for these layers~\cite{courbariaux2016binarized,liu2018bi,cai2017deep}. Second, they often explicitly omit quantization considerations of batch normalization and use standard full-precision computation as performed during training~\cite{cai2017deep,zhou2016dorefa}. Since batch normalization is essential to the convergence of low-precision CNNs, this omission makes it difficult to efficiently implement many low-precision approaches. In this work, as discussed in Section~\ref{sec:training-quantization}, we handle both of these issues by (1) quantizing the weights and data in all layers (including the first and last layers) under a single quantization scheme and by (2) including batch normalization quantization in the training graph (depicted in Figure~\ref{fig:quantization}) so that it adds zero overhead during inference.

\subsection{Weight Pruning}
It is well known in the literature that the majority of weights in a CNN (up to 90\% for large models such as VGG-16) can be set to zero (pruned) without having a significant impact on the classification accuracy~\cite{han2015deep}. The resulting pruned network may have sparsely distributed weights with an irregular sparsity structure, which is generally difficult to implement efficiently using conventional hardware such as GPUs. This has led to subsequent methods that propose structured pruning techniques which will result in models with nonzero weights densely distributed~\cite{wen2016learning,narang2017block,gray2017blocksparse,huang2017condensenet,he2017channel,luo2017thinet}. While these methods allow more efficient CPU and GPU implementations, they appear unable to achieve the same level of reduction in model size that unstructured pruning can achieve\footnote{For instance, in Table 4 of~\cite{wen2016learning}, the highest accuracy model relative to the number of nonzero weights is achieved using unstructured pruning.}.

\begin{figure}
    \centering
    \includegraphics[width=0.9\columnwidth]{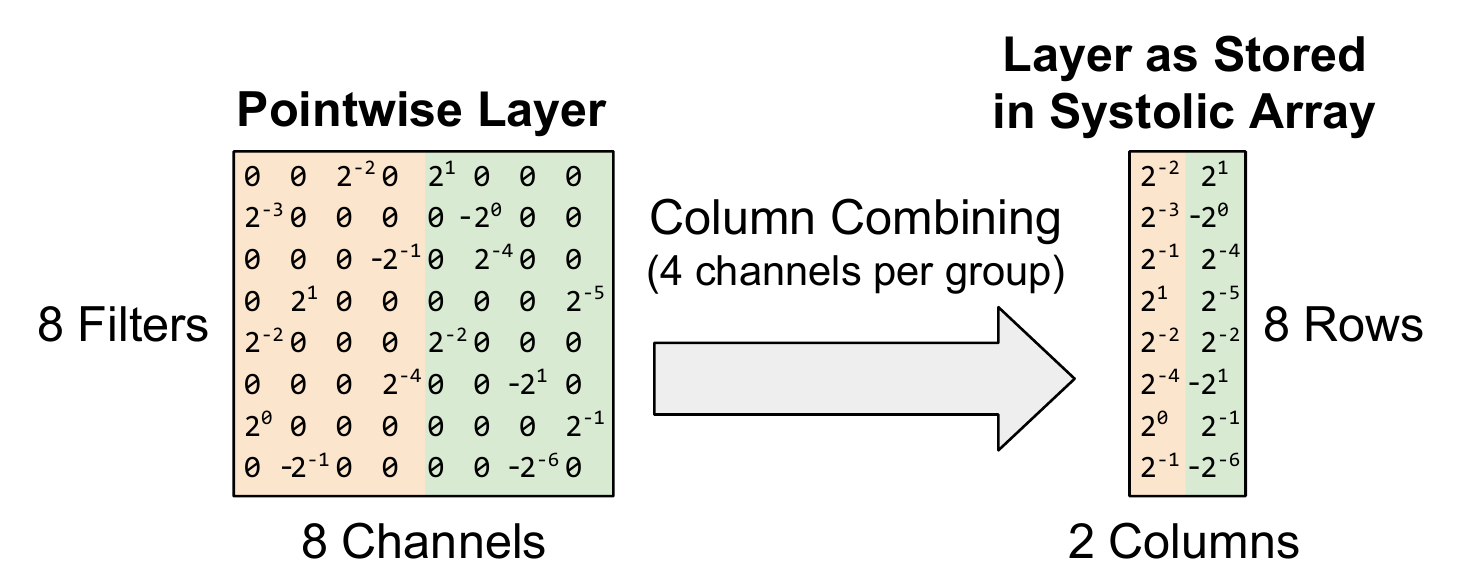}
    \caption{A pointwise convolution layer (left) with four channels per group resulting from weight pruning training for column combining~\cite{kungpacking18}. After combining columns in the filter matrix (left), each group of four channels (shown in cream and green) are reduced into a single column (right). Note that during column combing, for each group, all entries in each row are removed (pruned) but one with the largest magnitude.}
    \label{fig:combine-training}
\end{figure}

Unlike previous work, column combining is a new pruning method which allows for sparse CNN layers, but requires that the remaining sparse weights can be packed into a denser format when deployed in hardware~\cite{kungpacking18}. In our proposed training pipeline, we use column combining in addition to weight and data quantization as discussed in the previous section, in order to achieve efficient sparse CNN inference. Figure~\ref{fig:combine-training} shows how a sparse pointwise convolution layer with power of two weights is converted into a denser format by removing all but the largest nonzero entry in each row across the combined channels when stored in a systolic array. In this example, column combining reduces the width of the small layer by factor of 4$\times$ from 8 to 2. In Section~\ref{sec:fpga-inference}, we describe bit-serial design for efficient hardware implementation of this packed format shown on the right side of Figure~\ref{fig:combine-training}.

\section{Streamlined CNN Architecture}
\label{sec:training}

In this section, we first provide an overview of our streamlined CNN structure in Section~\ref{sec:cnn-structure}, targeted for our FPGA implementation reported in this paper. Then, we outline the various design choices to improve the utilization and efficiency of the FPGA system. Specifically, we employ a quantization-aware training graph, including quantized batch normalization (Section~\ref{sec:training-quantization}) and an input reshaping operation to improve the utilization of our systolic array for the first convolution layer (Section~\ref{sec:reshape-input}).  

\subsection{Proposed Streamlined CNN Architecture}
\label{sec:cnn-structure}

\begin{figure}
    \centering
    \includegraphics[width=0.65\columnwidth]{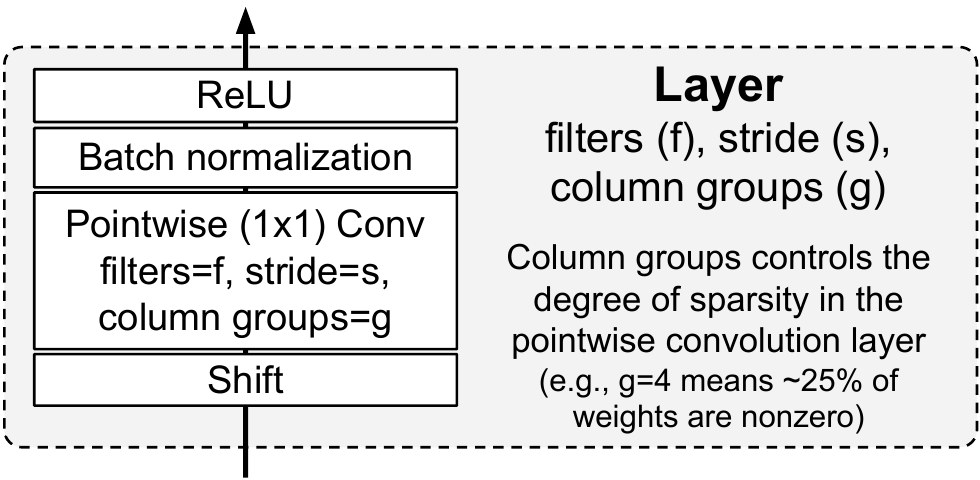}
    \caption{Each layer of the evaluation CNN models in this paper consists of a shift operation~\cite{wu2017shift}, pointwise (1x1) convolution, batch normalization and ReLU activation. A layer is parameterized with a number of filters (f), a stride (s), and column groups (g) for column combining in packing a sparse convolutional layer~\cite{kungpacking18}.}
    \label{fig:cnn-layer}
\end{figure}

Our objective in designing a CNN architecture is to achieve high classification accuracy using a simplified structure across all CNN layers which can be mapped efficiently onto a systolic array. Figure~\ref{fig:cnn-layer} shows the structure of each convolutional layer in our network. To achieve similar performance to standard 3$\times$3 convolution using only pointwise (1$\times$1) convolution, every layer begins with a channel shift operation as described in~\cite{wu2017shift}. The output of the \textit{shift operation} is then applied to a sparse pointwise convolution layer, followed by batch normalization and rectified linear unit (ReLU). During training, the weights in the pointwise convolution layer are pruned with column combining using the column groups parameter (g) as in~\cite{kungpacking18}. For the earlier convolution layers in a network which have fewer weights, a column group size of 2 is used, which reduces the number of nonzero weights by roughly 50\%. For the latter CNN layers, which are larger and have higher redundancy, a group size of 8 is used which reduces the number of nonzero weights by approximately 87.5\%. Each layer is progressively pruned over the course of training, such that after training they will reach their target sparsity set by the column groups for the layer.

\begin{figure}
    \centering
    \includegraphics[width=\columnwidth]{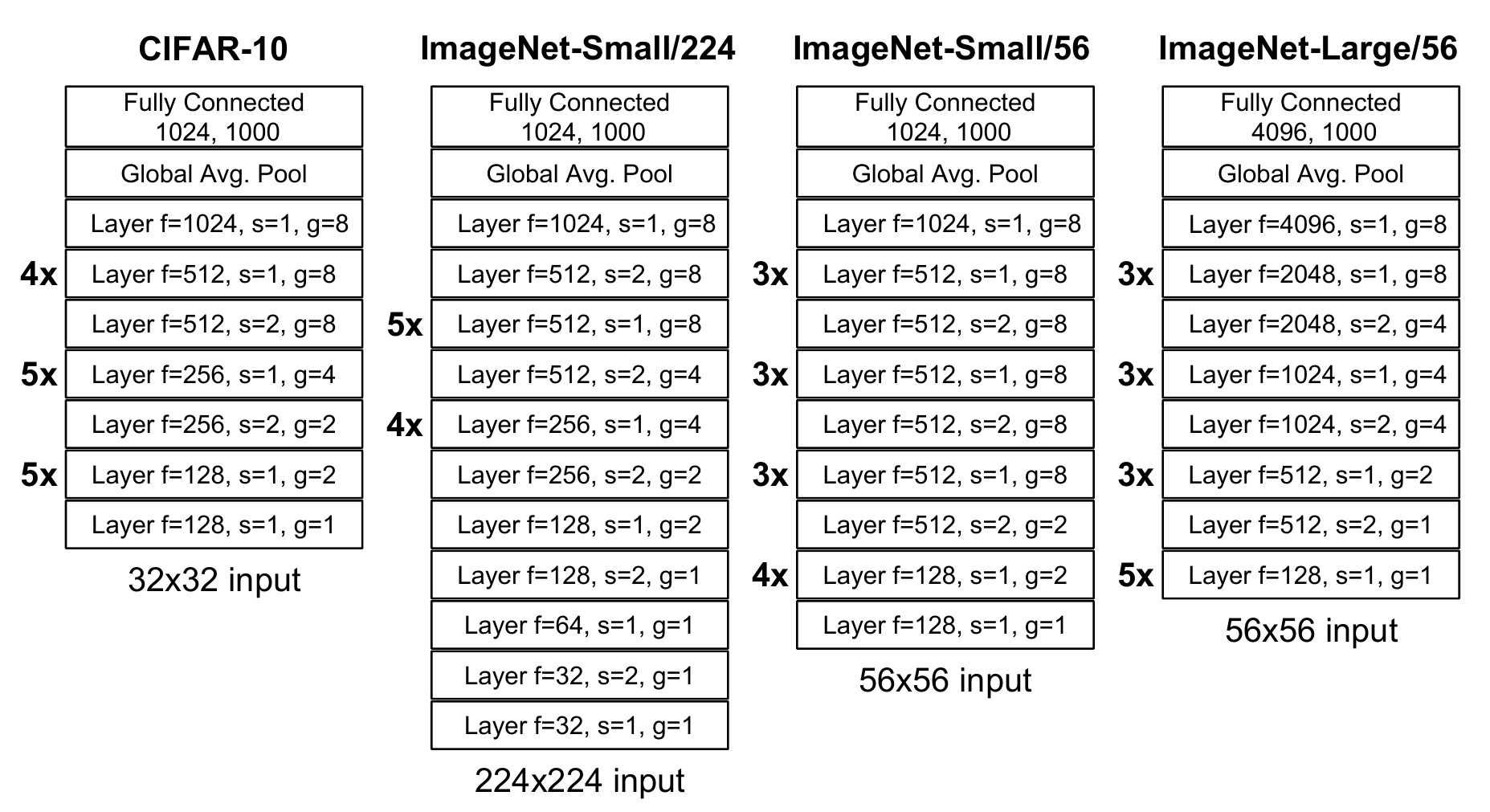}
    \caption{The evaluation models for the CIFAR-10 and ImageNet datasets. Each network consists of 19 layers, where each layer has a structure shown in Figure~\ref{fig:cnn-layer}, with the first layer not including a shift component. Downsampling is performed using strided convolution denoted by layers with a stride of 2 (s=2).}
    \label{fig:cnns}
\end{figure}

Figure~\ref{fig:cnns} shows the evaluation models for the proposed streamlined CNN structure for CIFAR-10~\cite{krizhevsky2014cifar} and ImageNet \cite{deng2009imagenet} datasets. As discussed in Section~\ref{sec:cnn-background}, we have chosen to keep the network relatively shallow (19 layers) and without any residual or concatenative connections. In Section~\ref{sec:eval}, we show that this streamlined structure can achieve competitive Top-1 ImageNet classification accuracy with low latency and high energy efficiency. We evaluate ImageNet using three settings: ImageNet-Small/224, ImageNet-Small/56, and ImageNet-Large/56, where 224 and 56 refer to the width and height of the input image after the prepossessing stage. The small models have 1.5M weights and the large model has 8.5M weights after training. These evaluation model were chosen to evaluate the importance of model size and the spatial input size on classification accuracy, latency, and throughput. Additionally, as described in Section~\ref{sec:reshape-input}, for the settings with (56$\times$56) input size, we use an \textit{reshaping operation} to increase the number of input channels from 3 (for RGB images) to 48 to the systolic array for low-latency input.
 
\subsection{Quantization-aware Training}
\label{sec:training-quantization}

\begin{figure}
    \centering
    \includegraphics[width=0.7\columnwidth]{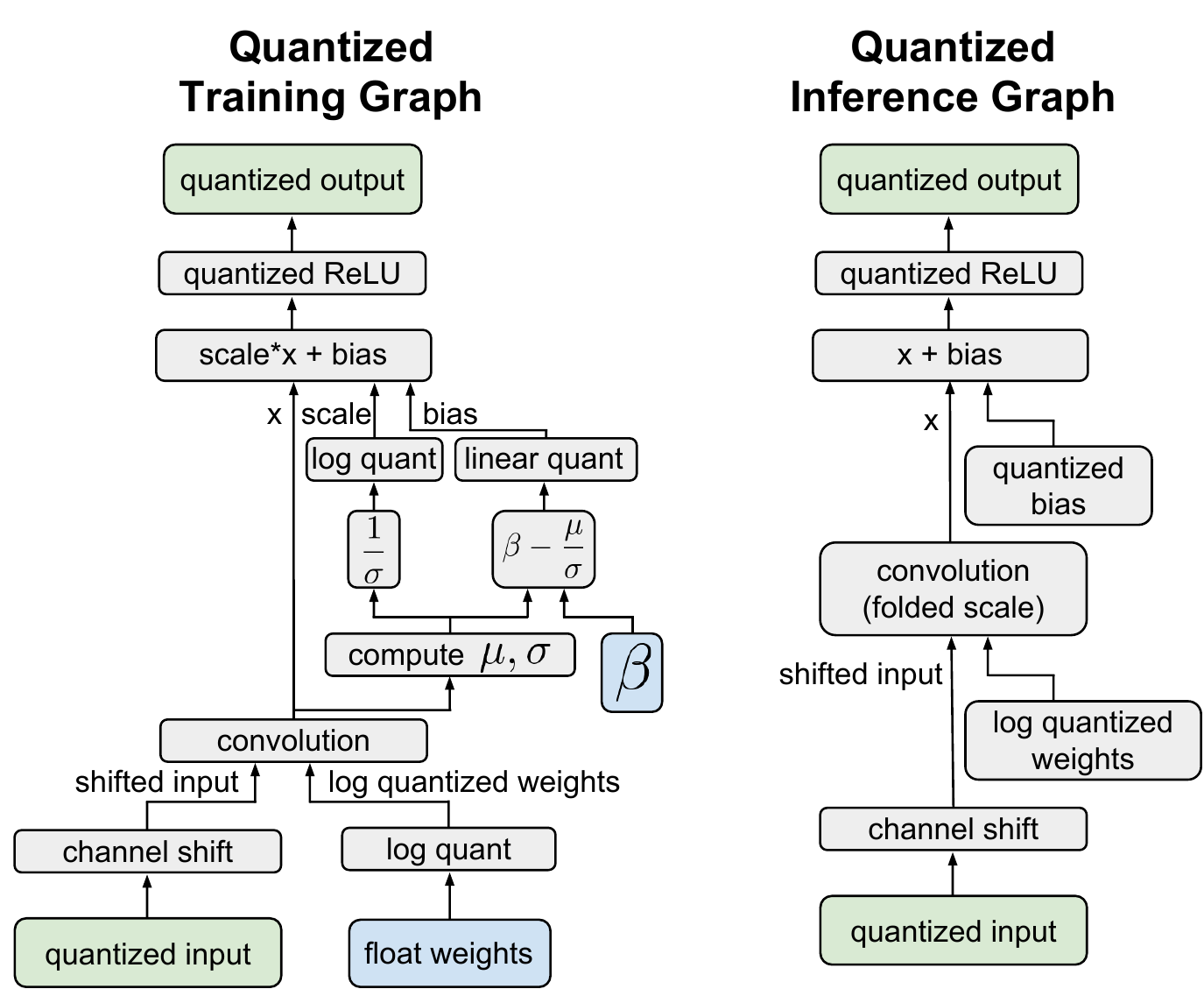}
    \caption{The quantized training graph (left) performs both linear quantization for input data and power of two quantization (log quantization) for weight and batch normalization parameters during training. The inference graph (right) uses the quantized version of the full-precision weights learned during training and therefore does not require any floating-point operations.}
    \label{fig:quantization}
\end{figure}

In order to achieve high classification accuracy using power of two weights, we add quantization operations to the CNN training graph in order to match the fixed-point weights and data used at inference. Figure~\ref{fig:quantization} shows the training and inference graphs for a single layer in the CNN shown in Figure~\ref{fig:cnn-layer}. As discussed in Section~\ref{sec:quantization}, this approach of injecting quantization into the training graph is known in the literature and has mainly been used to train binary and ternary networks~\cite{courbariaux2015binaryconnect,zhou2016dorefa}. In our training graph, we use log quantization for the weights, which quantizes an underlying full-precision weight (shown in blue) to the nearest power of two. During training, backpropagation uses full-precision gradients to update the full-precision weight in each layer.

Additionally, we perform quantization of the batch normalization operations which follow each convolutional layer. For higher precision weights (\eg~8-bit weights), these batch normalization parameters can be folded directly into the weights and bias terms of the preceding convolution layer after training, so that they have no additional overhead~\cite{krishnamoorthi2018quantizing}. However, for lower precision weights (such as binary or power of two weights), this folding processes introduces significant quantization error, leading to a notable drop in classification accuracy. For this reason, prior works using low-precision weights employ full-precision batch normalization incurring the corresponding full-precision computation (\eg~\cite{zhou2016dorefa}). For our proposed bit-serial architecture, these full-precision batch normalization operations would introduce significant overhead and break our objective of multiplication-free inference. Therefore, as shown in Figure~\ref{fig:quantization}, we include quantization of the batch normalization parameters in our training graph. Applying log quantization on the batch normalization scale parameters allows them to be folded into the log quantized weights without introducing any quantization error.

Batch normalization is defined as
\begin{equation*}
\label{equ:standardbn}
x_{bn} = \gamma (\frac{x-\mu}{\sigma}) + \beta
\end{equation*}
where $\mu$ and $\sigma$ are the mean and standard deviation of each mini batch during training and average running statistics during inference. $\gamma$ and $\beta$ are learnable parameters which are introduced to improve the representation power of the network. When followed by ReLU, as is the case in our CNN, the effects of the learnable scale parameter $\gamma$ can be captured in the following convolution layer and can therefore be omitted by setting gamma as 1~\cite{tfbn}.  We then factor $\mu$, $\sigma$, and $\beta$ into a scale and bias term as follows 
\begin{equation*}
\label{equ:foldbn}
    x_{bn} = \underbrace{\frac{1}{\sigma}}_{\text{scale}}x + \underbrace{\beta - \frac{\mu}{\sigma}}_{\text{bias}}
\end{equation*}

After applying quantized batch normalization to the output from the preceding convolution layer, a non-linear activation function (ReLU) is performed, which sets all negative values to 0. Additionally, it applies 8-bit linear quantization on the data so that it matches the fixed-point computation at inference. The inference graph of Figure~\ref{fig:quantization} shows how computation is performed on the FPGA during inference. The log quantized batch normalization scale factor is folded into the log quantized weights in the preceding convolution layer. Only arithmetic shift and fixed-point addition operations are performed during inference.

\subsection{Input Reshaping to Improve Utilization}
\label{sec:reshape-input}

For CNNs trained on ImageNet, the first convolution layer represents 10-15\% of the total computation performed due to the large spatial size of the input image (3$\times$224$\times$224). However, as recently discussed by Xilinx~\cite{xilinxtech18}, the computation in this layer does not map well onto systolic architectures, because the input image only offers three input channels, meaning that the majority of the array's input bandwidth may not be utilized. To address this imbalance, Xilinx proposes to use two systolic arrays, one systolic array specifically designated to the first layer and the other systolic array used for the remaining convolution layers in the CNN.

\begin{figure}
    \centering
    \includegraphics[width=0.8\columnwidth]{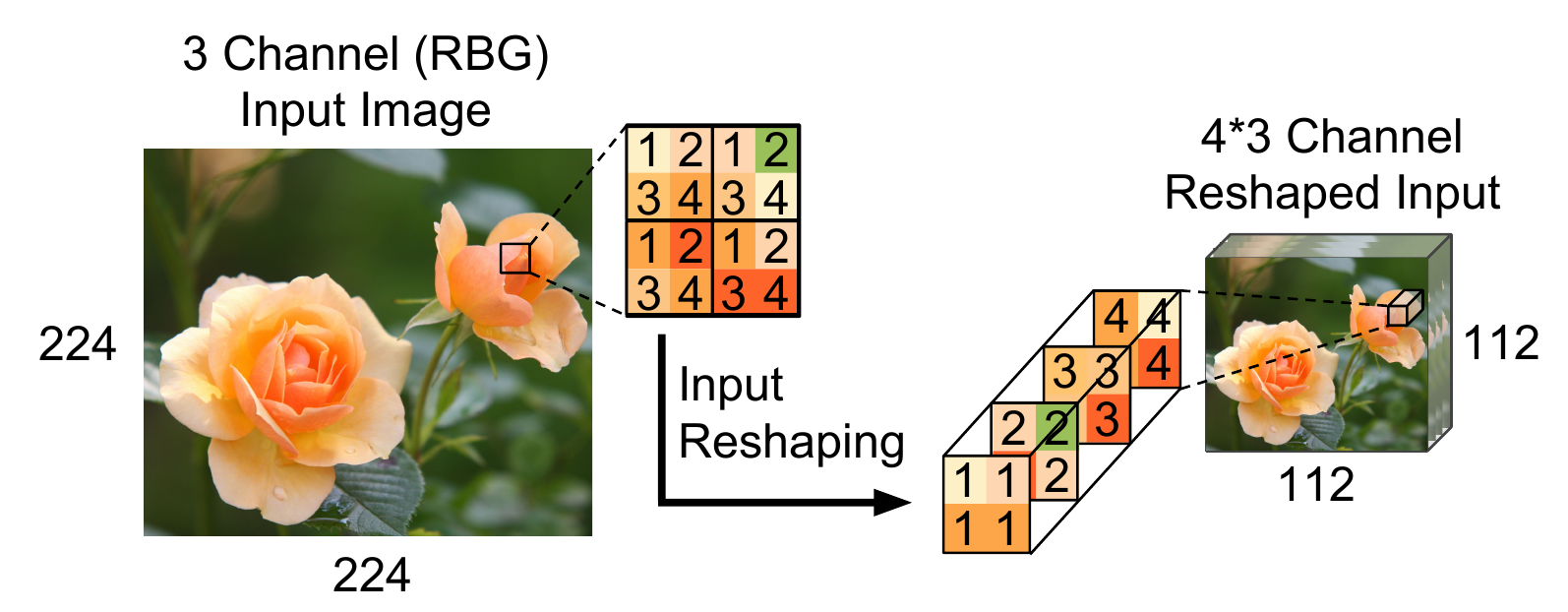}
    \caption{Reshaping the input data by decreasing the spatial size and increasing the number of channels in order to improve utilization of the systolic array in processing the first layer.}
    \label{fig:reshape-input}
\end{figure}

In this paper, rather than adding a systolic array for the input layer, we reshape the input image to the CNN to increase the utilization of our single systolic array for the first convolutional layer. Figure~\ref{fig:reshape-input} shows the input reshaping operation for an RGB image with 3 channels and 224$\times$224 pixels per channel. Each 2$\times$2 block of pixels is divided into four groups (denoted 1, 2, 3, and 4) with 1 pixel per group. The pixels in the same group across all 2$\times$2 blocks are then placed into a new set of RGB channels (4 groups, each with RGB channels, leading to 12 channels total). Each of these channels has 112$\times$112 pixels, which is one quarter of the original input image. In Section~\ref{sec:eval}, we evaluate the ImageNet-Small/56 and ImageNet-Large/56 networks with an even more aggressive reshaping operation, where we use 16 groups to convert the 3$\times$224$\times$224 input image into a 48$\times$56$\times$56 input for the CNN.

\section{Configuration and Instructions}
\label{sec:cnn-conversion}

In this section, we show how our trained CNN described in Section~\ref{sec:training} is coded for efficient configuration of systolic array on FPGA (Section~\ref{sec:coding}). We then explain how the weights in each layer are divided into tiles which fit in a given systolic array and the corresponding instructions for each tile which run on the FPGA (Section~\ref{sec:inst-generation}).  

\subsection{Coding Sparse CNN Layers for FPGA}
\label{sec:coding}

\begin{figure}
    \centering
    \includegraphics[width=0.9\columnwidth]{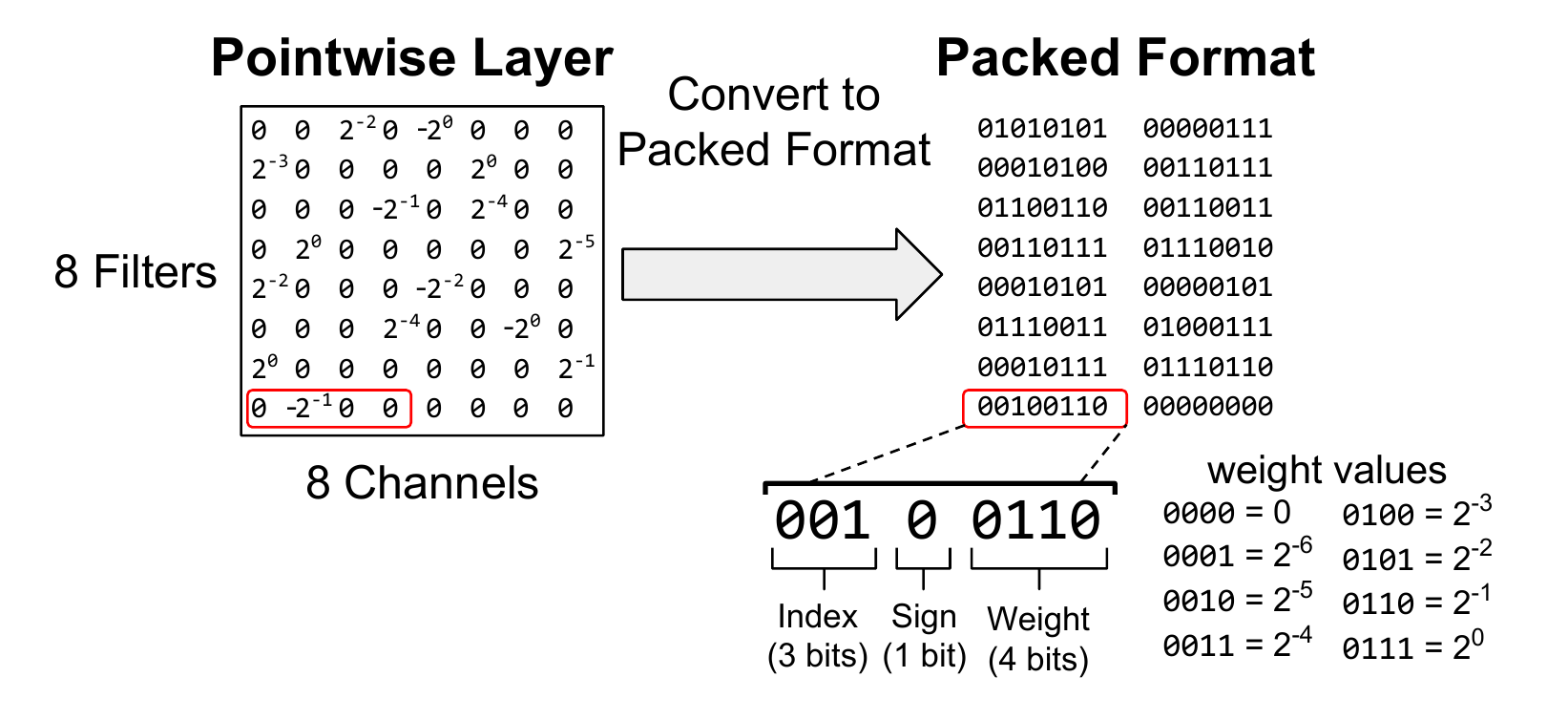}
    \caption{A sparse pointwise layer with power of two weights (left) is converted into a packed representation for efficient store on FPGA (right), where each group of combining channels (4 in this example) produces 1 8-bit encoding per filter (row).}
    \label{fig:packing-layer}
\end{figure}

After training is complete, each convolution layer will have reached a target sparsity set by the column group parameter for the layer as described in Section~\ref{sec:cnn-structure}. The left side of Figure~\ref{fig:packing-layer} illustrates the weights of a pointwise convolution layer after training with 8 filters, 8 channels, and column groups of size 4. For this layer, each group of  4 channels will be combined into a single column in the systolic array on the FPGA, as there is only one nonzero entry per filter (row) in each group. The remaining nonzero weights are power of two due to the weight quantization scheme discussed in Section~\ref{sec:training-quantization}.

To illustrate the coding procedure, we have outlined 4 weights in red in the pointwise layer shown in Figure~\ref{fig:packing-layer} which will be converted into an 8-bit representation in the packed format. The second element in this group is the nonzero weight $-2^{-1}$. The first 3 bits in the encoding store the index of the nonzero weight which is 1 (\texttt{001}) corresponding to the second element in the group. Note that for larger layers, we combine up to 8 channels, requiring a 3-bit index. The remaining 5 bits are a sign bit and 4 bits to indicate the power of two weight, which is \texttt{00110} for $-2^{-1}$. As depicted in Figure~\ref{fig:packing-layer}, the power of two weights are ordered from smallest to largest (\eg~$2^{-6}$ is \texttt{0001} and $2^0$ is \texttt{0111}). The value \texttt{0000} is used to represent $0$. In summary, to configure each systolic cell, only 8 bits are required.

\subsection{Instructions for FPGA}
\label{sec:inst-generation}

CNN inference is composed of a series of matrix-matrix multiplications, one per convolution layer, between the data which is input to a layer and the learned weights of a layer. When using a single systolic array to perform the matrix multiplications for all layers, generated instructions will carry out a relatively straightforward process of alternatively loading weights into the systolic array and performing matrix multiplication between the data and loaded weights, in sequential order of the CNN layers. However, when a weight matrix is larger than the fixed size of the systolic array, it must be partitioned into smaller \textit{tiles}, where each tile can fit into the systolic array. Then, a pair of weight loading and matrix multiplication instructions are scheduled for each tile. In this paper, we use column combining to dramatically reduce the number of columns for inference (\eg~from 512 channels to 64 columns in the systolic array via 8-way combining) and therefore require no tiling to partition channels of each convolutional layer of our evaluation CNNs tiling.

Figure~\ref{fig:tiling} shows how inference is performed across two layers (layer L and layer L + 1) using a single systolic array. First, a load weights instruction is used to load the 128 filters by 128 channels weight matrix for layer L. Then, matrix multiplication is performed between the loaded weights and the previous layer output (layer L - 1) by passing the data into the systolic array. This matrix multiplication generates the layer L output which will be used for layer L + 1. Since the layer L + 1 weight matrix of 256$\times$128 is larger than the systolic array of 128$\times$128, it is partitioned into two tiles. The first tile in layer L + 1 is then loaded into the systolic array and is multiplied with the layer L output, which generates half the output for layer L + 1. The second tile in layer L + 1 is then processed in the same manner as the first tile. A total of six instructions are used in total, one pair of weight load and matrix multiply instructions for layer L and two pairs of instructions for layer L + 1.  

\begin{figure}
    \centering
    \includegraphics[width=\columnwidth]{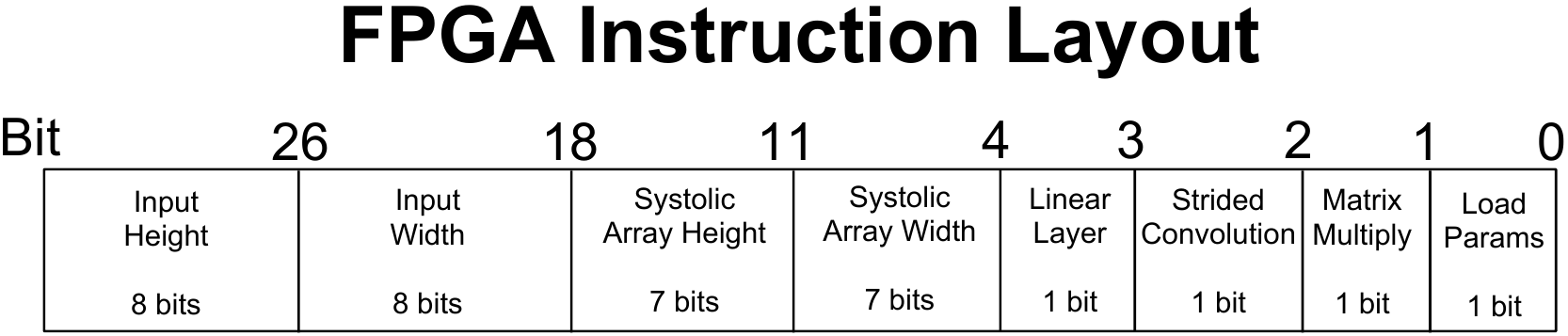}
    \caption{The FPGA instruction layout for a 128x128 systolic array on the FPGA.} 
    \label{fig:instruction-layout}
\end{figure}

Figure~\ref{fig:instruction-layout} shows the FPGA instruction layout for the systolic array architecture described in Section~\ref{sec:fpga-inference}. A load weight instruction is indicated when the first bit is set, with the systolic array width and height fields controlling the size of the tile being loaded into the array. A matrix multiply instruction is indicated when the second bit is set. The height of the data matrix to be multiplied with the loaded weights is set by the input width and height fields (\eg~56$\times$56 for the first layer). 

\section{FPGA Design}
\label{sec:fpga-inference}
In this section, we provide a detailed description of our FPGA design for sparse CNN inference with power of two weights. Our FPGA implementation is written in Verilog and is available at \url{https://goo.gl/8i9aJp}.

\subsection{System Description}
\label{sec:systolic-array-system}

\begin{figure}
    \centering
    \includegraphics[width=\columnwidth]{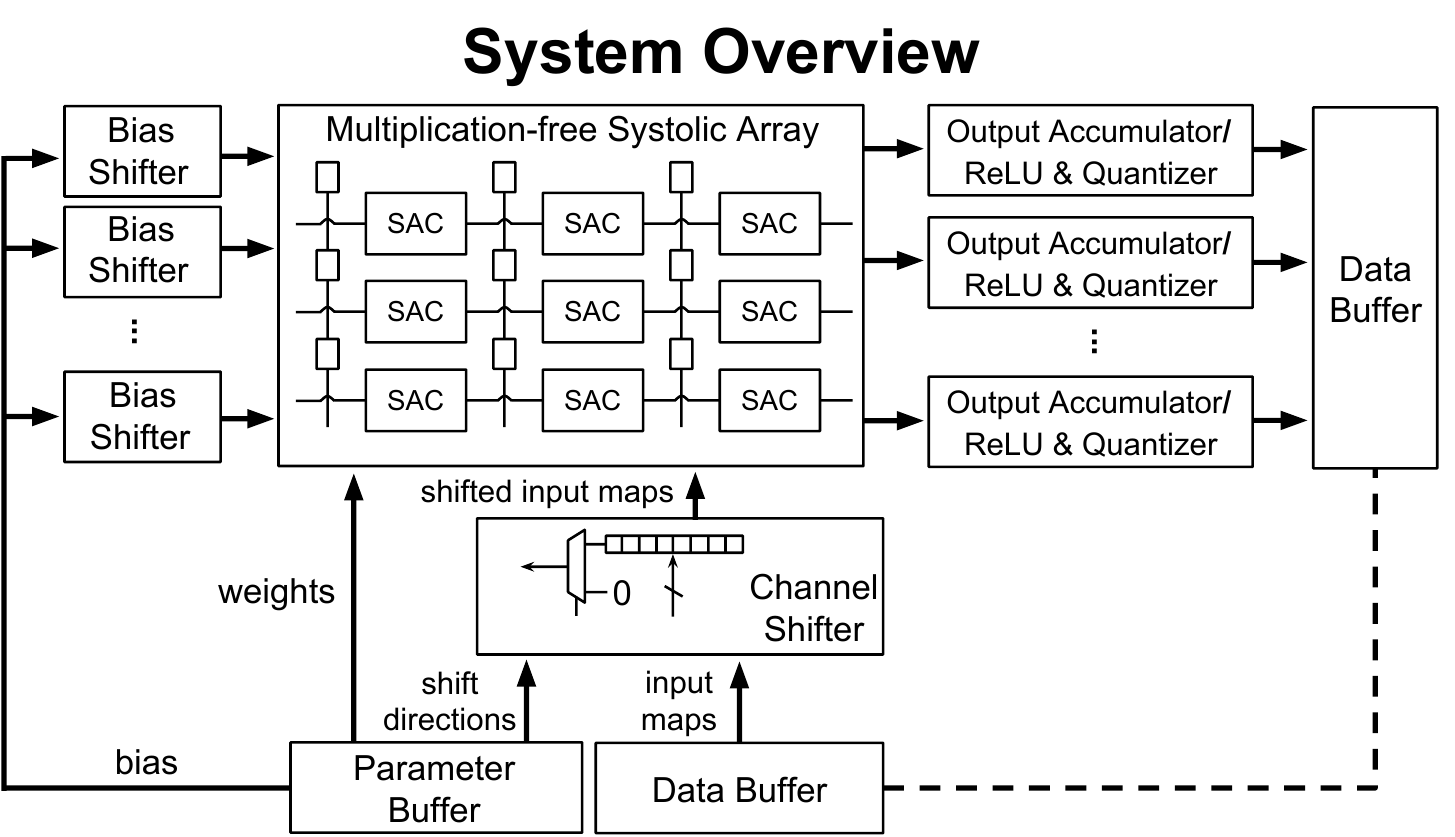}
    \caption{System design as implemented on an FPGA.}
    \label{fig:system-architecture}
\end{figure}

Figure~\ref{fig:system-architecture} shows an overview of the CNN inference system as implemented on an FPGA. The parameter buffer stores the filter weights, biases, and shift directions for the channel shift operation~\cite{wu2017shift}. During a load weight instruction, filter weights are loaded into the systolic array (Section~\ref{sec:mx-design}) and filter bias and channel shift directions are loaded into the bias and the channel shifters, respectively. During a matrix multiplication instruction, input data is loaded from the data buffer into the channel shifters, which perform the shift operations before sending the data to the systolic array in a bit-serial fashion. Each column in the systolic array takes input data from multiple input channels to support column combining shown in Figure~\ref{fig:combine-training}. Each Selector-Accumulator (SAC) cell (Figure~\ref{fig:bitserial-mac}) within a column of the systolic array takes in the multiple input channels at different power of two shift offsets to select both the channel index and power of two weight index for the cell corresponding to the packed format in Figure~\ref{fig:packing-layer}. 

The output from each row of the systolic array is passed to the ReLU $\&$ Quantization block (Section~\ref{sec:relu_quant}) before storing the results back to the data buffer. Output data stored in the data buffer for the previous layer is the input data to the next layer. The output accumulator (Section~\ref{sec:bias_and_accmulator}) is used only in the final (fully connected) layer to reduce the feature map for each class to a single number used for prediction. The parameters for the next tile are loaded from the off-chip DRAM (not shown in Figure~\ref{fig:system-architecture}) to the parameter buffer as matrix multiplication is performed on the systolic array for the current tile. During inference, all intermediate results are stored in on-chip RAM in the data buffer.

\subsection{Selector-Accumulator (SAC) for Multiplication-free Systolic Array Design}
\label{sec:mx-design}

In this section, we describe our Selector-Accumulator (SAC) design for a multiplication-free systolic array for sparse matrix multiplication. We choose a bit-serial design for efficient multiplexing of multiple data streams into a single column of the array to support column combining \cite{kungpacking18}. In the layout of the multiplication-free systolic array (shown in Figure~\ref{fig:system-architecture}), each column in the array takes up to eight input channels (to support column combining) into the register chain for the column in a bit-serial fashion. Each cycle, input data is shifted up to the next register in the chain. This register chain serves the standard purpose in a systolic array of propagating data through the cells in the column. However, when using power of two weights, it can serve an additional purpose of power of two weight multiplication, as each increasing position in the register chain corresponds to the input data being multiplied by an increasing power of two weight. In our design, we utilize this observation of the dual purpose nature of the register chain when using power of two weights to design an efficient systolic cell.

\begin{figure}
    \centering
    \includegraphics[width=\columnwidth]{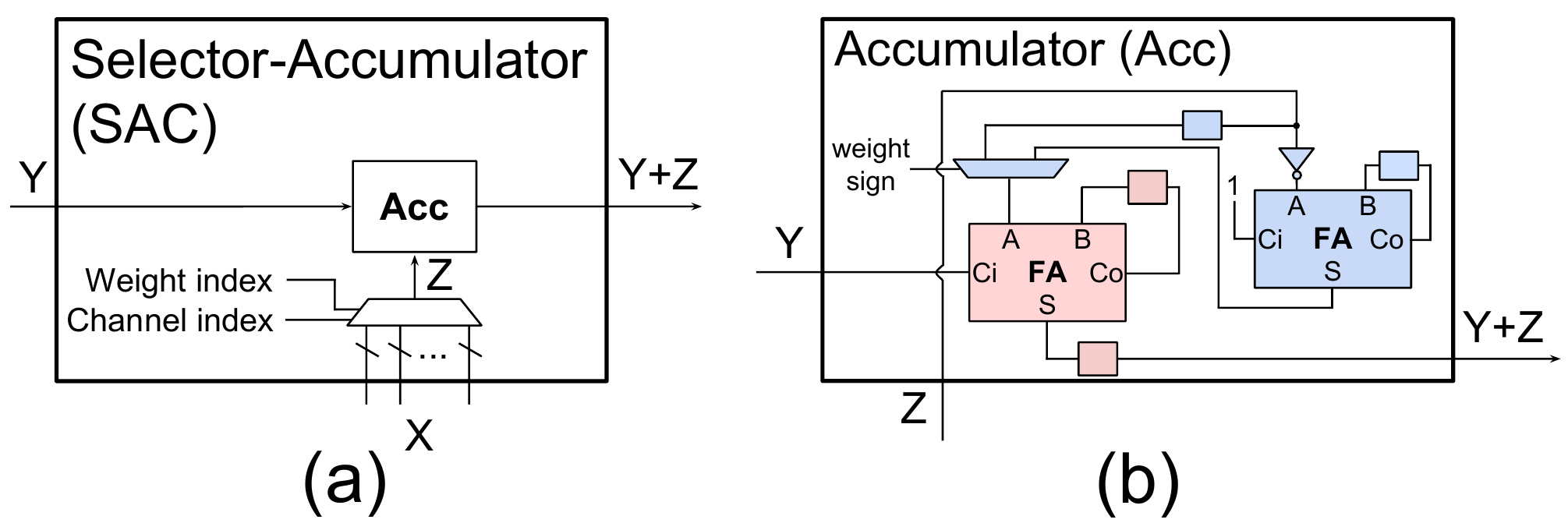}
    \caption{Bit-serial Selector-Accumulator (SAC).}
    \label{fig:bitserial-mac}
\end{figure}

Figure~\ref{fig:bitserial-mac}a shows the \textit{selector-accumulator} (SAC) cells which takes the input data from multiple points on the register chain and selects the point corresponding to the power of two weight stored in cell using a multiplexer. Additionally, it uses a channel index, also stored in the cell, to determine the position of the weight in the original sparse filter matrix (see Figure~\ref{fig:packing-layer} for details on the indexing scheme). The selected element is then passed into the bit-serial accumulator shown in Figure~\ref{fig:bitserial-mac}b. The blue logic elements in the accumulator negate the product Y based on the sign of the power of two weight and add the result to the bit-serial accumulator (pink full-adder) before passing the result to the SAC to the right.

Compared with a 8-bit multiplier-accumulator (MAC) which requires 8 1-bit full adders for multiplication, the SAC requires only a multiplexer to select the power of two shift offset. We have done performance comparisons using Xilinx Vivado design suite. We observed that compared to a traditional 8-bit MAC, SAC
substantially reduces required LUTs (4.85$\times$), FFs (3.54$\times$), and power
consumption (2.48$\times$), as shown in Table~\ref{tab:mac-sac-comp}. As we discuss in Section~\ref{sec:fpga-eval}, this dramatically reduces the hardware cost of each systolic cell and allows for substantially larger systolic array to be implemented on the FPGA compared to standard 8-bit MAC cells.

\begin{table}
\caption{Comparison of FPGA resources and power for a 64$\times$64 systolic array implemented with MAC and SAC.}
\label{tab:mac-sac-comp}
\begin{adjustbox}{width=0.7\columnwidth,center}
\begin{tabular}{l | c | c | c}
      & 64$\times$64 MAC & 64$\times$64 SAC & MAC / SAC \\ \hline
LUT   & 212388    & 43776     & 4.85$\times$     \\
FF    & 192293    & 54330     & 3.54$\times$     \\
Power & 4.21W     & 1.7W      & 2.48$\times$  \\  
\end{tabular}
\end{adjustbox}
\end{table}

\begin{figure}
    \centering
    \includegraphics[width=0.8\columnwidth]{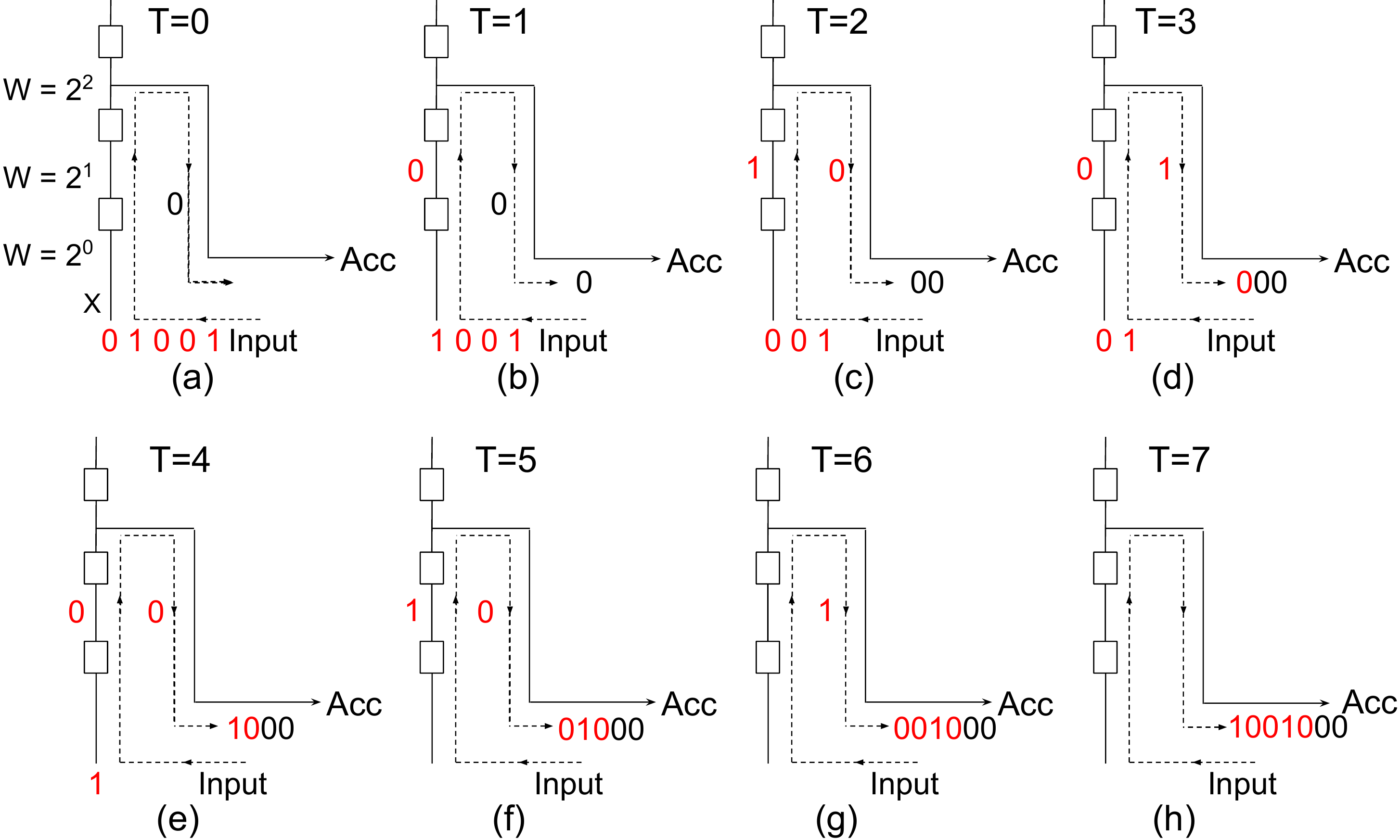}
    \caption{An example of sending the shifted version of input stream to the SAC cell.}
    \label{fig:shift_register_example}
\end{figure}

Figure~\ref{fig:shift_register_example} shows an example of how a register chain is used in generating a shifted version of the input data 10010 (red) with the shift amount corresponding to the power of two weight associated with the cell over time steps T = 0, 1, 2, etc. As depicted in Figure~\ref{fig:shift_register_example}~(a), (b) and (c), suppose that the SAC requires a shifted version of the original input with two pending zeros in the beginning (filter weight is four). Then the Accumulator (Acc) will grab the input data stream at the second register in the register chain, so the first two bits sent to the Acc are zeros (black). After 4 additional cycles, the Acc receives an input of 1001000, which is four times of the original input 10010.

\begin{figure}
\centering
\begin{minipage}[t]{0.6\columnwidth}
  \vspace{0pt}
  \centering
  \includegraphics[width=0.9\columnwidth]{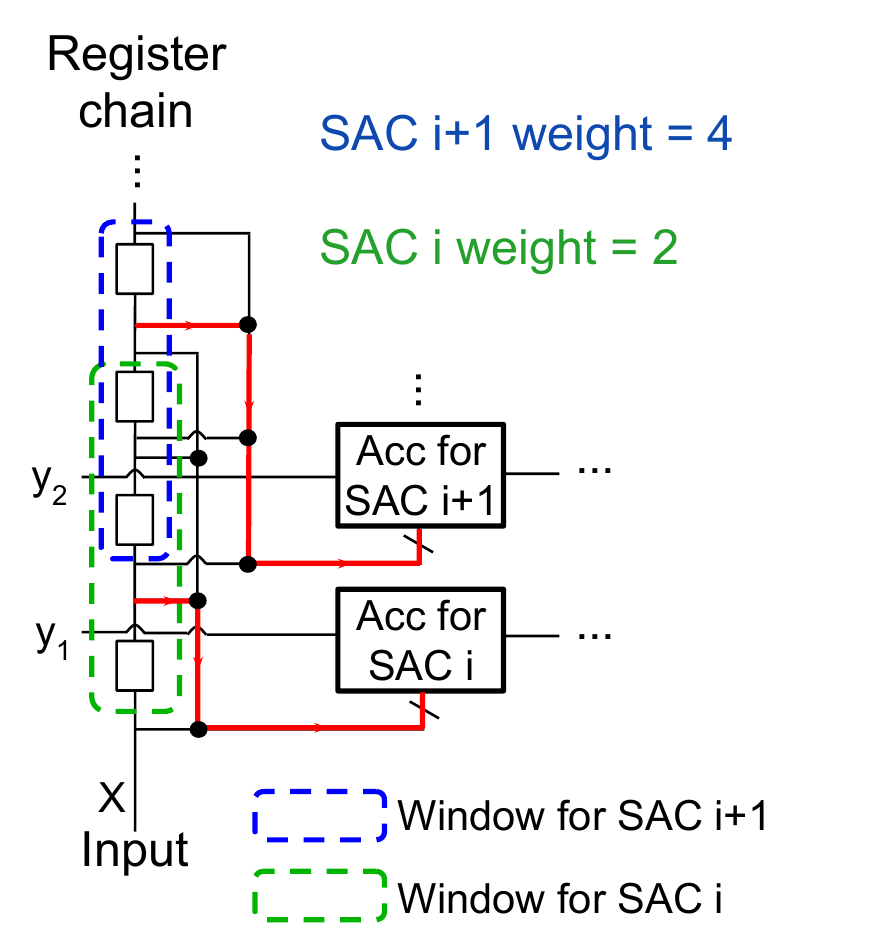}
  {\phantomsubcaption\label{fig:register_chain}}
\end{minipage}%
\begin{minipage}[t]{0.4\columnwidth}
  \vspace{0pt}
  \centering
  \includegraphics[width=\columnwidth]{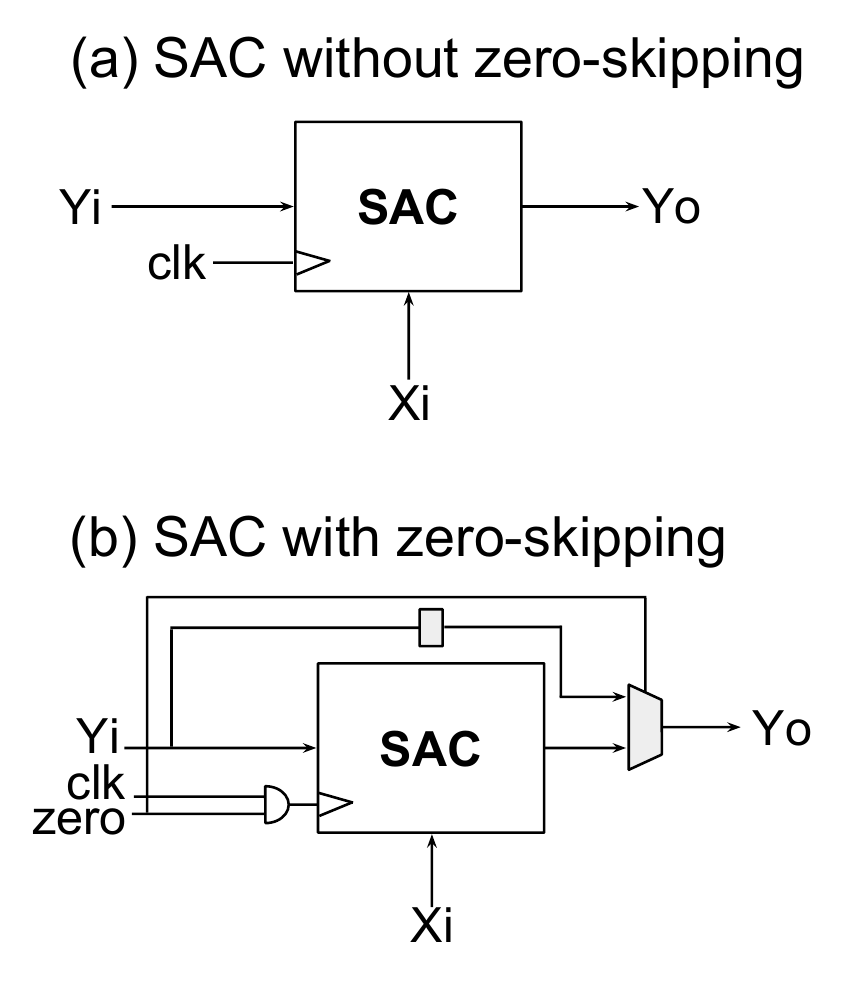}
  {\phantomsubcaption\label{fig:zero-skip}}
\end{minipage}%
\caption{(a) Register chain with per-cell window for power of two weights for two adjacent cells on a column of the systolic array. The red lines show the position where the shifted versions of the original input are grabbed from the register chain. (b) SAC without and with zero-skipping.}
\label{fig:shift-relu}
\end{figure}

Figure~\ref{fig:register_chain} shows how the register chain can be shared across two consecutive SAC cells in one column of systolic array. Suppose each of two SAC cells may require any one of the three shifted versions of the original input (corresponding to three possible powers of 2 weights). Then this leads to use of two windows with span of three on the register chain (shown in green and blue in Figure~\ref{fig:register_chain}). The red lines in the figures show the positions where the SAC cells grab the shifted versions of the original input from the register chain. The register chain is used for two purposes: (1) shifts the input data upwards to all the SAC cells in the same column and (2) generates the shifted versions of the input data for the power of two multiplication.

\subsection{Energy-efficient SAC with Zero-Skipping}
\label{sec:zero-skipping}

 Each SAC can be turned off to save power when either the weight or the input to the SAC is zero. The structure of a SAC with and without zero-skipping mechanism are shown in Figure~\ref{fig:zero-skip}. For the SAC with zero-skipping, the zero signal is set when either the input or weight is 0 and is used as the enable signal for the gated clock. When the zero signal is set due to the current input being 0, the accumulation $Y_i$ bypasses the SAC and is forwarded directly to the next SAC on the row in the systolic array. Note that due to ReLU, approximately half of the data elements are zero, meaning that the SAC will be disabled roughly half of the time. When the weight for the SAC is 0, then the SAC will be disabled for the entire matrix multiplication. In Section~\ref{sec:zeroskip-eval}, we show that this zero-skipping mechanism reduces power by roughly 30\%. 
 
\subsection{Design of ReLU and Quantization Block}
\label{sec:relu_quant}
As mentioned in Section~\ref{sec:quantization}, we use an 8-bit fixed point representation for the input data to each layer. Therefore, the quantization process must convert the higher precision (32-bit) accumulator output from the systolic array back into an 8-bit range to be used as input to the next layer. In this paper, since the fixed-point scale factor is shared across all layers, this quantization step simplifies to extracting an 8-bit range from the 32-bit accumulator. This quantization step can be fused with the ReLU activation function, by setting negative accumulator outputs to 0. 

\begin{figure}
    \centering
    \includegraphics[width=0.8\columnwidth]{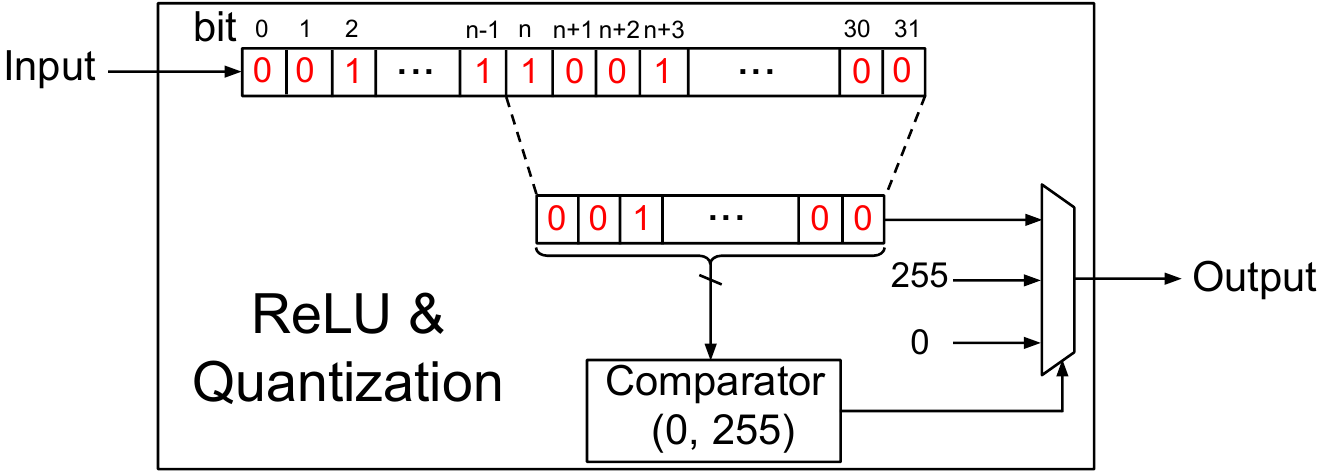}
    \caption{Design of ReLU \& Quantization block.}
    \label{fig:relu_quantizer}
\end{figure}

Figure~\ref{fig:relu_quantizer} shows the architecture of the ReLU \& Quantization block. A register array is used to collect the 32-bit output from the systolic array. The 32-bit result is shifted by the smallest representable power of two weight (\eg~$2^{-6}$ as shown in Figure~\ref{fig:packing-layer}) and passed to the comparator. The comparator generates the indicator bit for the multiplexer, which clips the result between $(0,255)$ before storing it back to the buffer.

\subsection{Design of Output Accumulator}
\label{sec:bias_and_accmulator}
Given the output channels produced by the final convolutional layer, average pooling is used to reduce the spatial components of each channel to a single averaged value. For our single systolic array implementation, we fold this pooling operation into the weights of fully connected layer. Let $x^{k}_{i}$ and $\bar{x}^{k} = \frac{\sum_{i=1}^{R} x^{k}_{i}}{R}$ denote the i-th element of the input map $k$ and the average of the input channel $k$, where $R$ is total number of elements in each channel. Denote $\vec{x} = \{\bar{x}^{1},\bar{x}^{2},...,\bar{x}^{M}\}$ as the vector of channel averages, where $M$ is the total number of input channels. We have the following derivation for the output $\vec{y}$ of the fully connected layer:
\begin{align}
    \vec{y} &= W\vec{x} 
            = W\frac{\sum_{i=1}^{R}\vec{x_{i}}}{R} 
            = \sum_{i=1}^{R} \frac{W}{R}\vec{x_{i}} \label{eqn:d5}
\end{align}
where $\vec{x_{i}} = \{x_{i}^{1},x_{i}^{2},...,x_{i}^{M}\}$ and $W$ is the weight matrix of the fully connected layer. From equation~\ref{eqn:d5}, we notice that $\frac{W}{R}\vec{x_{i}}$ can be computed by carrying out the matrix multiplication between $\frac{W}{R}$ and $X = \{x^{k}_{i}\}$ with the systolic array, and $\vec{y}$ can be computed by summing up all the $\frac{W}{R}\vec{x_{i}}$. 

The output accumulator is used to calculate the sum of $\frac{W}{R}\vec{x_{i}}$. The 32-bit output stream from a row in the systolic array enters the output accumulator in a bit-serial fashion. This stream is stalled in a register array until the final bit arrives. The resulting 32-bit output is added to the next 32-bit output stream. We use DSPs to carry out part of these 32-bit additions. 

\section{Evaluation}
\label{sec:eval}

In this section, we first briefly reiterate the key contributions of our proposed architecture from the perspectives of both the CNN training and hardware and tie each contribution to the corresponding evaluation section (Section~\ref{sec:design-summary}). Then, we evaluate the performance of our FPGA implementation against state-of-the-art FPGA accelerators on the ImageNet dataset (Section~\ref{sec:fpga-eval}). Next, we measure the impact of zero skipping on energy efficiency (Section~\ref{sec:zeroskip-eval}) for different sized systolic arrays. Finally, in Section~\ref{sec:cnn-eval}, we analyze the impact of our streamlined CNN structure and training procedure presented in Section~\ref{sec:training} on classification accuracy including input reshaping, using power of two weights, and the omission of residual connections from the CNN structure.

We focus on two primary performance metrics: (1) \textit{latency} from image input to classification output, and (2) \textit{energy efficiency}, which is the number of images the inference engine can process per joule. Note that the latter is also the number of images/sec (i.e., throughput) per watt. For high-throughput inference applications we may use multiple inference engines in parallel.  If these individual engines each offer low latency and high-energy efficient inference, then the
aggregate system will delivery high-throughput inferences per watt while meeting low inference latency requirements.

\subsection{Recap of Full-stack Optimization}
\label{sec:design-summary}

Full-stack optimization via training has enabled the following design advances which lead to our efficient FPGA implementation presented in Section~\ref{sec:fpga-inference}.

\begin{itemize}
    \item Using power of two for weights and the batch normalization scale parameters, outlined in Section~\ref{sec:quantization}, for all layers in the CNN (including the fully connected layer). This allows for a simplified design, where a single sparse multiplication-free systolic array is used for all CNN layers. In Section~\ref{sec:quant-eval}, we discuss the impact of the proposed quantization scheme on classification accuracy.
    \item Zero-skipping of the quantized data (Section~\ref{sec:zero-skipping}). In Section~\ref{sec:zeroskip-eval}, we show that zero-skipping reduces the power consumption during matrix multiplication by roughly 30\%. 
    \item Packing sparse CNNs using column combining~\cite{kungpacking18} for efficient storage and use on FPGAs, which we describe in Section~\ref{sec:coding}. Our ImageNet-Small/56 evaluation model has only 1.5M power of two weights, which is 40$\times$ smaller than AlexNet and 92$\times$ smaller than VGG-16 (the two CNNs used by other FPGA designs).
    \item Using channel shifts~\cite{wu2017shift} to replace 3$\times$3 convolutions with 1$\times$1 convolutions. As with column combining, this reduces the number of model parameters. Additionally, it streamlines the design of the systolic array system, as 1$\times$1 reduces to matrix multiplication.
    \item Input reshaping (Section~\ref{sec:reshape-input}) to increase the bit-serial systolic array utilization and dramatically reduce the latency for the first convolution layer. In Section~\ref{sec:reshape-eval}, we show that input reshaping alleviates the accuracy loss when using a smaller spatial input size of 56$\times$56 instead of the conventional 224$\times$224.

\end{itemize}

\subsection{Comparing to Prior FPGA Accelerators}
\label{sec:fpga-eval}

We compare our 170 MHz FPGA design to several state-of-the-art FPGA accelerators on the ImageNet dataset in terms of top-1 classification accuracy, latency for a single input image, and energy efficiency when no batch processing is performed (\ie~batch size of 1). By choosing these metrics, we focus on real-time scenarios where input samples must be processed immediately to meet a hard time constraint. Our evaluation model is the ImageNet-Small/56 network shown in Figure~\ref{fig:cnns} with input reshaping to 48$\times$56$\times$56. Our FPGA can fit a systolic array of size 128 rows by 64 columns. Each of the columns can span up to 8 channels in convolution weight matrix,~\ie~when the column group parameter is set to 8, for a total of 512 channels.

Table~\ref{tab:overall} provides a comparison of our FPGA implementation with the other FPGA-based CNN accelerators. Our design achieves a per-image latency of 2.28 ms, which is among the lowest across all the designs. In addition, compared with some of the most recent works~\cite{zhang2018dnnbuilder,wang2018design}, our design outperforms them by 5.64$\times$ and 3.26$\times$, respectively, in term of energy efficiency. Additionally, compared to an implementation which achieves comparable low latency~\cite{li2016high}, our implementation has 9.29x higher energy efficiency. 

Our design achieves the highest energy efficiency among all these designs for several reasons. First, we use a highly efficient CNN structure (Section~\ref{sec:cnn-structure}) with only 1.5M weights (compared to 60M for AlexNet and 136M weights for VGG-16~\cite{ma2017optimizing,qiu2016going}). Our model in Table~\ref{tab:overall} is significantly smaller and all weights (including weights in batch normalization layers) are quantized to power of two numbers. Our accuracy is $50.84\%$ (about 2\% worse than nearest competitive designs~\cite{wang2018design} in terms of energy efficiency).
However, our implementation  has at least 3x  higher  energy efficiency. 
Second, our proposed power of two quantization (Section~\ref{sec:quantization}) enables the use of a multiplication-free systolic array (Section~\ref{sec:systolic-array-system}), where each cell contains only a selector and two full adders (see Figure~\ref{fig:bitserial-mac}) which are more efficient compared with~\cite{ma2017optimizing} and have simpler structure compared with~\cite{shen2017maximizing}. This allows for a large systolic array (128$\times$64) to fit on the FPGA, thereby reducing the number of tiles required to perform inference for each sample. Moreover, by using column combining~\cite{kungpacking18} we can pack sparse CNN layers for efficient systolic array implementation with high hardware utilization~\cite{xiao2017exploring}. Additionally, DSPs are used in the Output Accumulator (Section~\ref{sec:bias_and_accmulator}) only for a single fully connected layer and are turned off for the rest of the layers. Finally, the zero-skipping mechanism, which we evaluate in more detail in Section~\ref{sec:zeroskip-eval}, further saves power by dynamically turning off systolic cells when the data entering a cell is zero. 

\begin{table}
\caption{Comparison with FPGA-based CNN accelerators.}
\centering
\begin{adjustbox}{width=\columnwidth,center}
\begin{tabular}{lcccccccc}
\hline
                            &        \cite{zhang2018dnnbuilder}            &    \cite{qiu2016going}                &     \cite{xiao2017exploring}                &         \cite{ma2017optimizing}          &        \cite{li2016high} &\cite{shen2017maximizing}   & \cite{wang2018design} & Ours                 \\ \hline
Xilinx FPGA Chip                   & VC706      &  ZC706      &  ZC706       & Arria-10  &  VC709    &   Virtex-7 & ZC706 &  VC707        \\

FF               & 51K(12\%)                & 127k(29\%)                & 96k(22\%)                 & -               & 262k(30\%)      &348k(40\%)         & 51k(12\%) & 201K(33\%)                \\
LUT               & 86k(39\%)                & 182k(83\%)                & 148k(68\%)                 & 161k(38\%)               & 273k(63\%)               & 236k(55\%) &86k(39\%) &239K(78\%)                  \\
DSP               & 808(90\%)                & 780(89\%)                & 725(80\%)                 & 1518(100\%)                & 2144(59\%)      &3177(88\%)         &808(90\%)& 112(4\%)                  \\

BRAM               & 303(56\%)                & 486(86\%)                & 901(82\%)                 & 1900(70\%)                & 1913(65\%)      &1436(49\%)         & 303(56\%) & 834(81\%)                  \\
Accuracy \small{(Top-1)}          & 53.30\%            & 64.64\%            & N/A             & N/A            & N/A    & 55.70\%     &52.60\% & 50.84\%        \\
Frequency \small{(MHz)}               & 200                & 150                & 100                 & 150               & 150              & 100 & 200 &170                  \\

Latency (ms)                 & 5.88               & 224                & 17.3                & 47.97             & 2.56          &11.7  & 5.84 & \textbf{2.28}            \\
Efficiency \small{(img./S/W)} & 23.6               & 0.46              & 6.13               & 0.98              & 12.93        &8.39  & 40.7 & \textbf{120.7}          \\

\hline
\end{tabular}
\label{tab:overall}
\end{adjustbox}
\end{table}

\subsection{Power Reduction by Zero Skipping}
\label{sec:zeroskip-eval}
In order to evaluate the power reduction due to zero skipping, we measure the power consumption of the FPGA during matrix multiplication under two settings. The ``Without Skipping'' setting uses inputs which are all nonzero, meaning that every cell will be active during matrix multiplication. The ``With Skipping'' setting uses inputs which are half zero, in order to approximate the output of ReLU, which sets roughly half of the elements to zero.

Table~\ref{table:zeroskip} shows the amount of power consumption for inference for the ``Without Skipping'' and ``With Skipping'' settings for three systolic arrays of increasing sizes. For all three systolic array sizes, we observe that ``With Skipping'' reduces the power consumption of matrix multiplication by roughly 30\%.

\begin{table}[h]
\centering
\caption{Power consumption comparison of zero-skipping.}
\begin{adjustbox}{width=0.6\columnwidth,height=0.08\columnwidth,center}
\begin{tabular}{l|c|c}
      & Without Skipping & With Skipping  \\ \hline
32$\times$64 &     1.0W        &        0.7W       \\ \hline
64$\times$64 &     1.7W        &        1.3W       \\
\hline
128$\times$64 &    3.0W        &       2.2W       \\
\end{tabular}
\end{adjustbox}
\label{table:zeroskip}
\end{table}

\subsection{Impact of Full-stack Training on Accuracy}
\label{sec:cnn-eval}
We now evaluate the impact of the modifications to both the CNN structure and training procedure as proposed in Section~\ref{sec:training} on classification accuracy.

\subsubsection{Impact of Input Reshaping}
\label{sec:reshape-eval}
In order to determine the effectiveness of the input reshaping operation described in Section~\ref{sec:reshape-input}, we compare models using the same spatial input size with and without reshaping (\eg~3$\times$56$\times$56 versus 48$\times$56$\times$56) and models with different spatial input size (\eg~3$\times$224$\times$224 versus 48$\times$56$\times$56). Additionally, we train a larger ImageNet model (ImageNet-Large/56) using input reshaping to see best accuracy that our proposed approach can achieve when used with a small spatial input size.

Table~\ref{table:input-reshape} shows the classification accuracy for the four evaluated network settings. First, we observe that the ImageNet-Small/56 with reshaping is able to achieve similar classification accuracy to the  ImageNet-Small/224 without reshaping, even with a 16$\times$ fewer pixels in each channel. This shows that input reshaping allows for input images with additional channels to negate the loss in accuracy due to the small spatial input size. Additionally, for the two ImageNet-Small/56 models (with and without reshaping), we see that input reshaping provides a substantial improvement of around 4\% accuracy. This is especially interesting considering these two networks have identical structures except for the initial layer (48 channels with input reshaping versus 3 channels without reshaping). Finally, the ImageNet-Large/56 model achieves an impressive 67.57\% which is only 2\% behind full-precision MobileNet using 224$\times$224 input. This shows that the proposed CNN structure and power of two quantization method can achieve high classification accuracy with reshaped input.
 
\begin{table}[]
\centering
\caption{Evaluating impact of input reshaping.}
\begin{adjustbox}{width=0.8\columnwidth,height=0.1\columnwidth,center}
\begin{tabular}{l|c|c}
Model & Input Reshaping & Accuracy (\%) \\ \hline
ImageNet-Small/224  & No    & 52.32 \\
ImageNet-Small/56   & No    & 46.92 \\
ImageNet-Small/56   & Yes   & 50.84 \\
ImageNet-Large/56   & Yes   & 67.57 \\  
\end{tabular}
\end{adjustbox}
\label{table:input-reshape}
\end{table}
 
\subsubsection{Impact of Power of Two Weight Quantization}
\label{sec:quant-eval}

While power of two weight quantization allow for an exceedingly efficient implementation, they introduce some loss in classification accuracy when compared against a full-precision version of the same network. Additionally, if these schemes are only evaluated on easier datasets (such as CIFAR-10), the reduction in accuracy can be understated when transition to harder datasets (such as ImageNet). Table~\ref{table:quant-loss} shows the classification accuracy for the CIFAR-10 and ImageNet-Small/56 models using full-precision and power of two weights. We see that while the gap between the CIFAR-10 models is only around 2.5\%, the gap for ImageNet is closer to 6\%. However, as we demonstrate in Section~\ref{sec:reshape-eval}, this reduction in classification accuracy can often be alleviated by increasing the model size.
 
\begin{table}[]
\centering
\caption{Comparing classification accuracy (\%) for full-precision and power of two weights for the CIFAR-10 and ImageNet-Small/56 models.}
\begin{tabular}{l|c|c}
               & CIFAR-10 & ImageNet-Small/56 \\ \hline
Full-Precision & 95.28    & 57.16  \\
Power of Two  & 92.80    & 50.84                 
\end{tabular}
\label{table:quant-loss}
\end{table}

\subsubsection{Impact of Removing Residual Connections}
\label{sec:residual-comp}
\begin{figure}
    \centering
    \includegraphics[width=0.65\columnwidth]{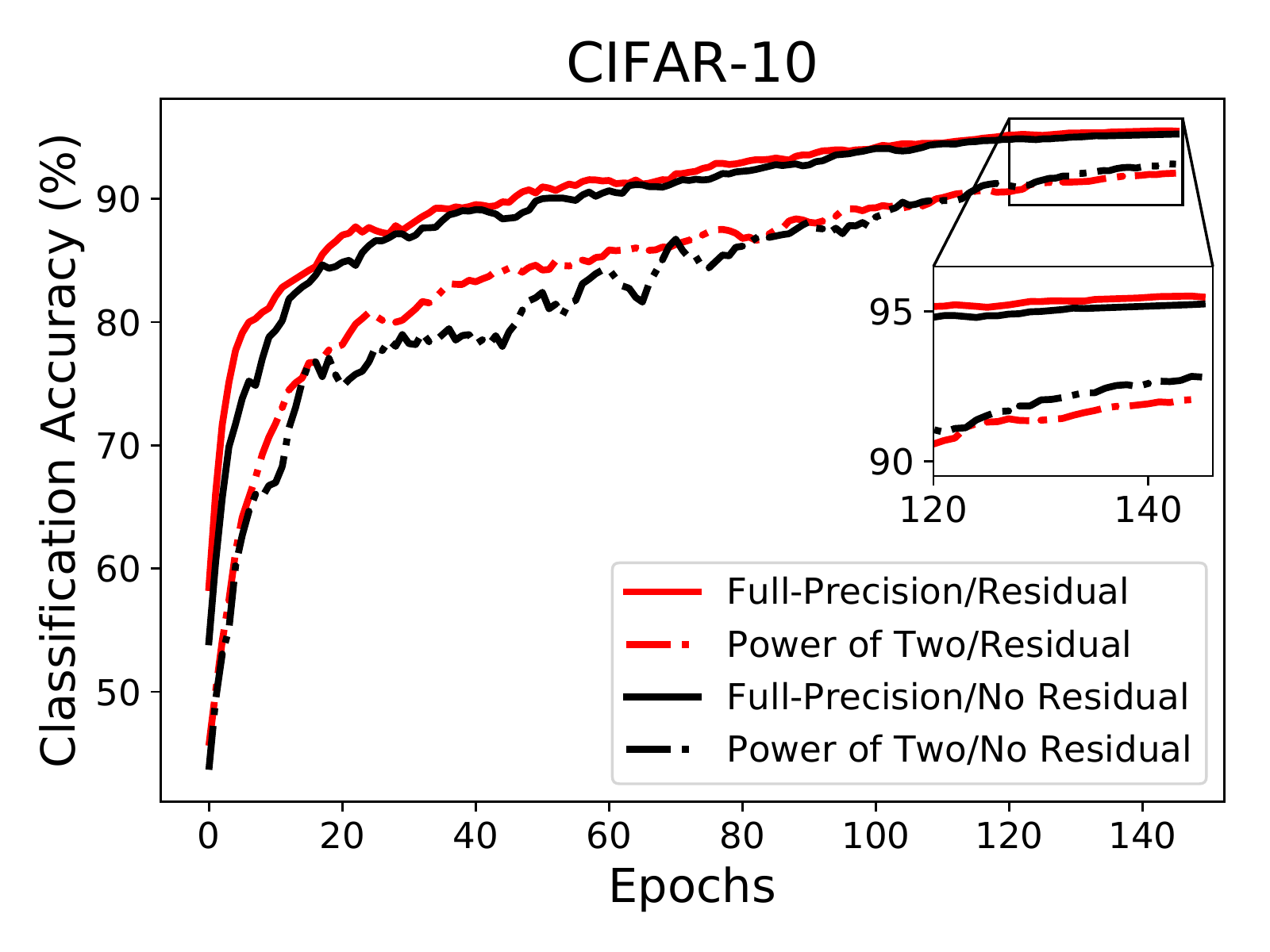}
    \caption{Classification accuracy over 150 epochs CIFAR-10 models trained with/without residual connections and with/without power of two quantization.}
    \label{fig:residual-comp}
\end{figure}

Figure~\ref{fig:residual-comp} shows the impact of residual connections by evaluating the CIFAR-10 network structure with and without residual connections. In order to ensure that there is not an unseen interaction between the power of two quantization and residual connections, we compare the impact of residual connections on networks with and without quantization. We see that, regardless of quantization, networks trained without residual connections achieve similar performance to the networks trained with residual connections. This shows that residual connections have minor impact on classification accuracy for the 19 layer networks as shown by He~\etal~in the original ResNet paper~\cite{he2016deep}.

\section{Conclusion}
\label{sec:conclusion}

In this paper, we propose using full-stack optimizations for accurate, low-latency and high energy-efficiency CNN inference. We demonstrate that designs ranging from CNN model training at a high level, to those of computing structures and FPGA implementation at a low level can all be optimized simultaneously to ensure they fit one another, thereby achieving high system performance. While cross-layer optimization is a known concept in the literature, to the best of our knowledge, the system reported in this paper is one of the most comprehensive realizations based on full-stack optimization for the design of deep learning implementations on a chip.

We describe implementation details of various optimization techniques, including (1) channel shifts instead of computationally more expensive 3$\times$3 convolutions, (2) packing sparse CNNs of irregular sparsity structure for efficient implementations on regular processor arrays, (3) quantizing data activations for power-saving with zero-skipping and efficient storage of intermediate data between layers, and (4) use of power of two weights and batch normalization for efficient computation.

Our Selector-Accumulator (SAC) design resulting from full-stack optimization with power of two weights represents an extremely efficient way of implementing MAC by selecting from a shift register rather than performing arithmetic operations. (It would be difficult to have a more efficient MAC design, short of analog implementations!) Given that MAC is the basic operation in the dot-product computation for matching data against filters, we believe our SAC result is significant.

\bibliographystyle{ACM-Reference-Format}
\bibliography{ref}

\end{document}